\documentclass[Afour,sageh,times]{sagej}

\usepackage{moreverb,url}
\usepackage[colorlinks,bookmarksopen,bookmarksnumbered,citecolor=red,urlcolor=red]{hyperref}
\usepackage{siunitx}
\usepackage{multirow}
\usepackage{makecell}
\usepackage{dirtree}
\usepackage{listings}
\usepackage{xcolor}
\usepackage{subcaption} % Required for creating the grid layout
\usepackage{graphicx}
\usepackage{etoolbox}
\usepackage{censor}

\newcommand\BibTeX{{\rmfamily B\kern-.05em \textsc{i\kern-.025em b}\kern-.08em
T\kern-.1667em\lower.7ex\hbox{E}\kern-.125emX}}

\def\volumeyear{2026}

\def\secref#1{Sec.~\ref{#1}}
\def\figref#1{Fig.~\ref{#1}}
\def\tabref#1{Tab.~\ref{#1}}
\def\eqref#1{Eq.~(\ref{#1})}

\newcommand{\method}{{SignScene}}
\newcommand{\dataset}{\textbf{SiGNgapore}}

\makeatletter
\DeclareRobustCommand\onedot{\futurelet\@let@token\@onedot}
\def\@onedot{\ifx\@let@token.\else.\null\fi\xspace}

\makeatother

\usepackage{pifont}% http://ctan.org/pkg/pifont
\newcommand{\cmark}{\ding{51}}%
\newcommand{\xmark}{\ding{55}}%

\colorlet{punct}{red!60!black}
\definecolor{background}{HTML}{EEEEEE}
\definecolor{delim}{RGB}{20,105,176}
\colorlet{numb}{magenta!60!black}

\lstdefinelanguage{json}{
    basicstyle=\normalfont\ttfamily,
    numberstyle=\scriptsize,
    stepnumber=1,
    numbersep=8pt,
    showstringspaces=false,
    breaklines=true,
    backgroundcolor=\color{background},
    literate=
     *{:}{{{\color{punct}{:}}}}{1}
      {,}{{{\color{punct}{,}}}}{1}
      {\{}{{{\color{delim}{\{}}}}{1}
      {\}}{{{\color{delim}{\}}}}}{1}
      {[}{{{\color{delim}{[}}}}{1}
      {]}{{{\color{delim}{]}}}}{1},
}

\begin{document}

\runninghead{Zimmerman et al.}

\title{\dataset{} - An Interactive Dataset for Sign-based Visual Navigation}

\author{Nicky Zimmerman\affilnum{1}, Joel Loo\affilnum{1}, Zishuo Wang\affilnum{1}  and David Hsu\affilnum{1}}

\affiliation{\affilnum{1}National University of Singapore, SG}
%\\
%\affilnum{2}SAGE Publications Ltd, UK}

\corrauth{Nicky Zimmerman, National University of Singapore
Smart Systems Institute,
3 Research Link,
117602, SG.}

\email{nicky.zi@nus.edu.sg}

\newcommand{\firstpagefigure}{%
  \begin{minipage}{\textwidth}
    \centering
    \includegraphics[width=0.95\linewidth]{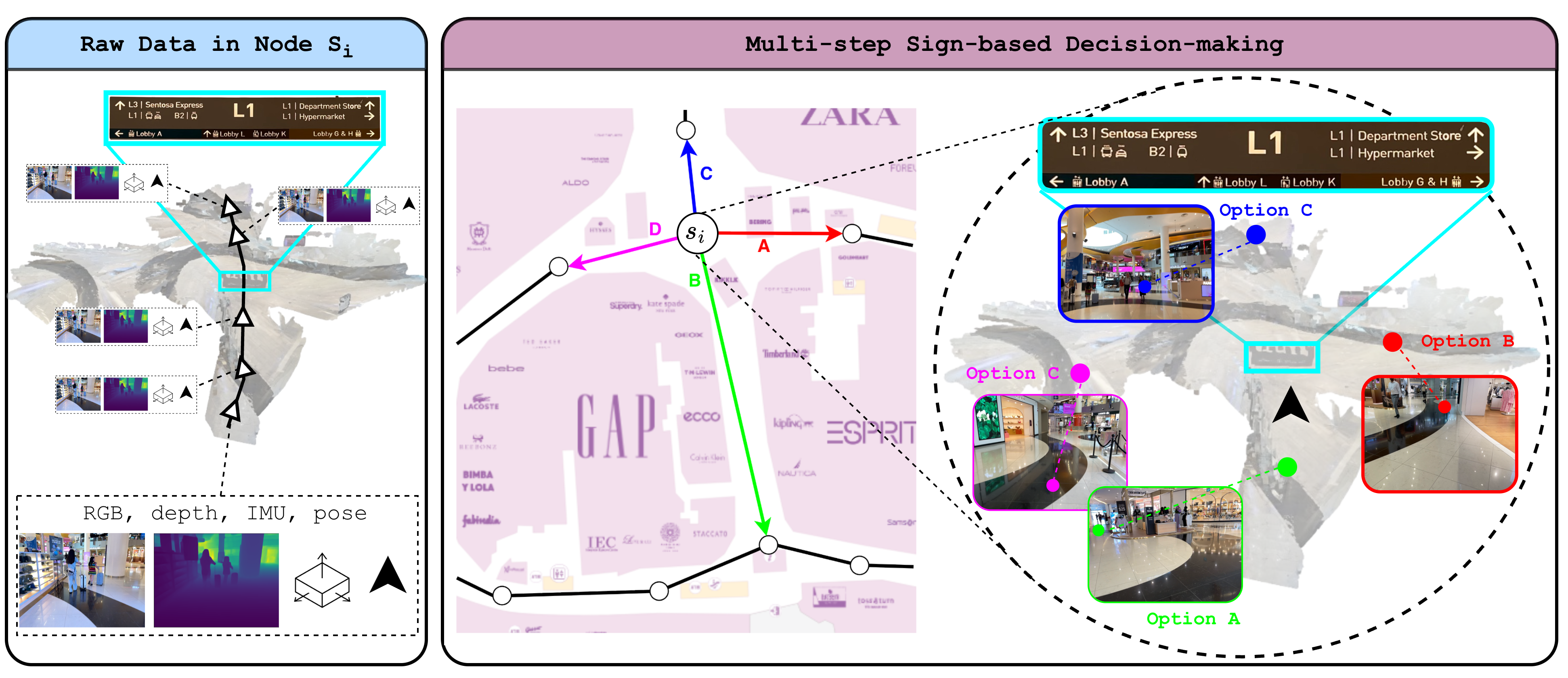}
    %{pics/Topograph.png}
    \captionof{figure}{%
      \textbf{Multistep decision making.} Our dataset aims at evaluating large-scale sign-based navigation. Left: Each scenario node $S_i$ includes a sequence of sensor measurements capturing a navigational sign and the scene around it. Right: Given a goal, defined by a mission, the agent should reason over the sign and ground the current navigational action from among multiple VQA options. Based on the chosen action, the agent will encounter the next scenario, $S_j$, according to the topological graph of each environment which connects decision-making points. }
    
    \label{fig:scenariodyn}
  \end{minipage}
  \par\vspace{1em}
}

\makeatletter
\patchcmd{\@maketitle}
  {\usebox\absbox}
  {\firstpagefigure\usebox\absbox}
  {}
  {\PackageError{firstpagefigure}
    {Could not find the abstract insertion point}
    {Check the definition of \string\@maketitle\space in sagej.cls}}
\makeatother

\begin{abstract}
% The future of autonomous robots in human-oriented environments depends on their ability to navigate in the absence of prebuilt maps; exploiting navigational aids, such as navigational signs, designed by humans for humans is critical for enabling autonomy. This manuscript introduces a unique dataset collected using a handheld device, at various public spaces across Singapore, representing environments that humans frequent daily, and focusing on sign-centric decision making for visual navigation. The dataset includes RGB, sparse depth, odometry and IMU measurements of scenarios centered around navigational signs and the complex environments where they are placed. Additionally, we provide 56 long-horizon navigation missions and over 450 sign-centric scenarios. All data is provided in a human-readable format, as well as utility scripts for conversion to ROS 2. Lastly, we provide venue maps and GPS aligned scene graphs of the test environments. We discuss the potential use cases for this dataset.

Robots navigating open-world, human-centered environments must look beyond prebuilt high-fidelity maps and leverage navigational aids embedded in their surroundings, particularly, \textit{signs}. To advance sign-guided visual navigation, we introduce \textit{SiGNgapore}, an interactive, multimodal dataset designed to benchmark sign-guided multi-step navigational decision-making. SiGNgapore covers diverse public spaces in Singapore—including shopping malls, public transportation hubs, and hospitals—featuring multi-floor, multi-building layouts and indoor-outdoor transitions. It consists of 56 long-horizon navigation missions and 456 scenes, each containing a navigational sign and its surrounding environment. 
We provide synchronized RGB, depth, IMU, and odometry data for each scene. We additionally provide the layout map and the GPS-aligned scene graph for selected mission sites.
Beyond navigational decision-making, SiGNgapore supports a variety of spatial-semantic reasoning tasks, including navigational sign parsing and visual sign grounding, with the ultimate aim of driving progress in mapless robot navigation.

\end{abstract}


\maketitle

\begin{figure*}
\centering
\includegraphics[width=0.95\textwidth]{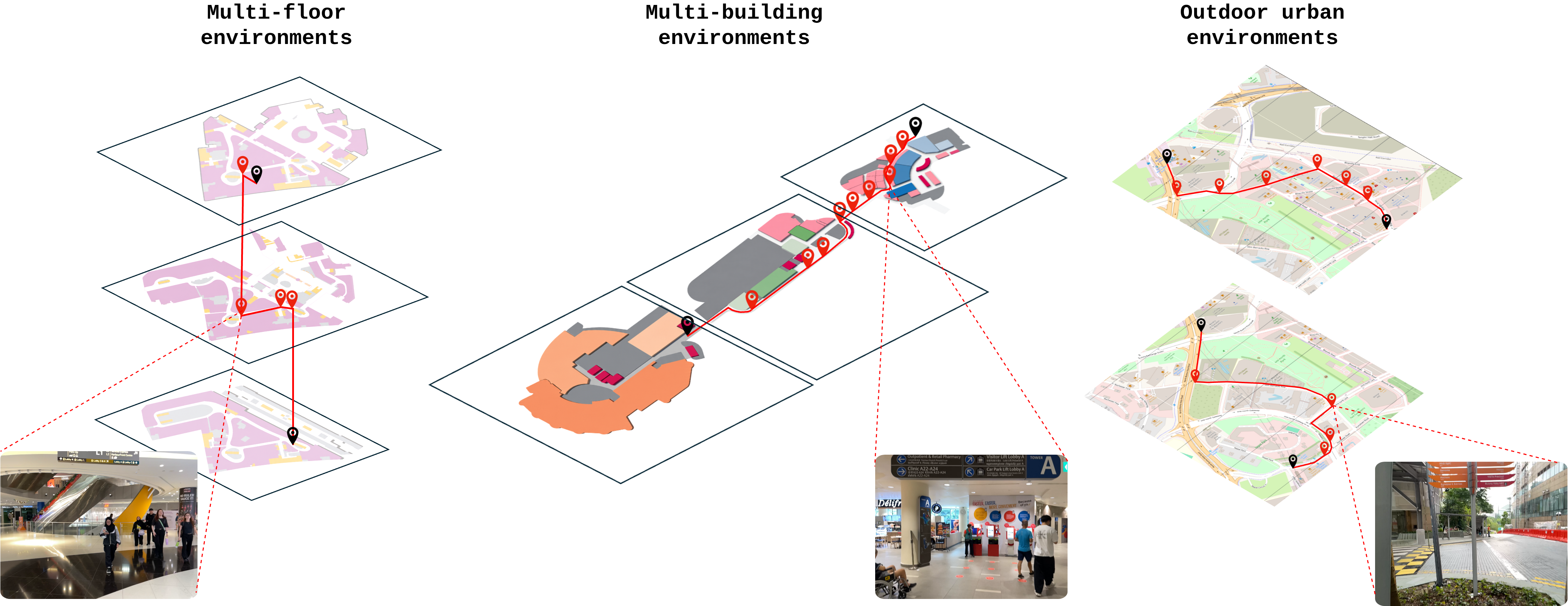}
%{pics/mission_diversity.png}
\caption{\textbf{Diversity of missions and environments in dataset.} We collate a range of missions across indoor and outdoor environments, including common public spaces from malls to hospitals to parks. Indoor missions extend across multiple floors and interconnected buildings.}
\label{fig:mission}
\end{figure*}

\section{Introduction}

Human-built spaces are often densely annotated with navigational signage to guide humans through unfamiliar environments without maps. However, robots at present make little use of these highly informative semantic cues, instead relying heavily on prior maps or other learned priors to guide them in navigation and search. To advance robots' capabilities in this respect, we introduce a dataset to benchmark decision-making in navigation based on signage. Our dataset collates a wide variety of ``missions'' across diverse public spaces in Singapore~(\figref{fig:mission}). Each ``mission'' is a multi-step navigation task, requiring an agent to accurately perceive and interpret information from signage, then select appropriate navigational decisions based on the interpreted information~(\figref{fig:scenariodyn}). We posit that a benchmark for this task can drive progress in mapless navigation using powerful semantic cues already embedded in everyday environments, empowering robots to navigate open-world, human-centric environments.

% The data should be placed in the context of current research making it clear which research field it applies to
% If applicable, a commentary on the data is valued. Discuss aspects of the data (or its interpretation) whose processing is within and outside the state of the art at the time of submission.
% The paper should be driven from start to end by the desire to make the data used by other authors. At the time of review, a link to the dataset’s webpage must be provided. This website can be password protected during the review process but the dataset must be publicly available after publication of the data paper. A permissive license that allows for free academic use must be attached to the dataset. If the authors’ submission is intended for an audience/use already addressed by an existing
% (published) Data Paper, then the authors should be explicit about what shortcomings are addressed by their submission
% \textcolor{red}{it's about decision making!!}
% This article describes a dataset collected by a hand-held device across multiple public spaces in Singapore. The platform is equipped with a selected sensors designed to enable resource-aware visual navigation research.

%Why it's important?

Mapless navigation is an increasingly critical capability for autonomous agents~\citep{anderson2018cvpr,ku2020emnlp, zhang2025rss}. In particular, there is growing interest in using navigational signs as cues for mapless navigation in human-oriented environments~\citep{agrawal2026icraws}, driven by the emergence of powerful Vision-Language Models (VLMs) that offer broad visual and linguistic understanding. Various approaches to sign-based navigation have been proposed~\citep{liang2020iros, chandaka2025corl, cao2026arxiv, chen2025ral}, but typically make strong simplifying assumptions about the structure and content of the navigational signs encountered. While several works seek to relax these assumptions to interpret and ground broader varieties of signs~\citep{agrawal2026icraws, zimmerman2026isrr}, the perceptual understanding of signs still remains a significant challenge even with existing VLMs~\citep{agrawal2026icraws}. More importantly, existing sign-based navigation approaches have yet to evaluate on multi-step, sequential decision-making tasks, where agents must sequentially interpret and act upon multiple navigational signs to reach a destination.

This motivates the need for a real-world dataset to benchmark sign-based visual navigation, particularly in the context of multi-step decision-making. At present, the only real-world sign-related dataset is provided by \cite{agrawal2026icraws}, and exclusively targets the narrower problem of perceiving and interpreting signs. While \cite{sun2026arxiv} and \cite{lin2025arxiv} design simulation environments that contain signs, these are limited to static, generic and limited environments that do not include realistic signage or environments, and do not fully reflect the difficulty of navigating in visually and topologically distinct human-oriented spaces. 

To this end, we propose \dataset{}, a dataset that evaluates sequential sign-based decision-making over long-horizon navigation missions. It comprises 456 distinct real-world scenarios, each capturing a navigational sign and its surrounding environment, which collectively underpin 56 long-horizon navigation missions that require agents to sequentially interpret and act upon multiple signs to reach a destination. Similar to NaviTrace~\cite{windecker2025arxiv}, \dataset{} takes a Visual Question Answering (VQA) format, to effectively evaluate decision-making capability on real-world data. In \dataset{}, the agent is required to reason on a sequence of observations collected around the navigational sign, capturing dynamics and 3D scene information. The dataset includes synchronized and timestamped range, visual, inertial and odometric measurements, collected in multiple public spaces in Singapore, such as malls, hospitals, campus buildings, transportation hubs and outdoor areas. As shown in \autoref{fig:mission}, these public spaces cover many of the environments that humans navigate daily, and enabling mapless navigation in such environments will greatly improve the autonomy of mobile agents. 
In addition to data from the hand-held device, multiple venue maps of the test site are included. The dataset goes beyond single-step visual sign grounding, by introducing long-horizon missions with multi-step decision making. 

Our dataset distinguishes itself from other datasets
described in the literature in several ways. Specifically, it
provides:
\begin{enumerate}
\item[(i)] Real-world sign-centric scenarios for visual sign grounding
\item[(ii)] Complex, topologically and visually distinct environments capturing the difficulty of reasoning about navigational actions in human-built spaces
\item[(iii)] Multi-step structure for mapless, sign-centric long-range navigation missions
\end{enumerate}
We evaluate the latest work on sign-based mapless navigation, \method{}~(\citet{zimmerman2026isrr}).
The dataset will be highly relevant to researchers working on embodied navigation, open world navigation, semantic scene understanding and cognition. 
% The remainder of this article describes the test environments and
% sensor setup, the format of the dataset, and the tools that are
% provided to process the data. 
%We also discuss some of the ways in which the dataset might be utilized in different branches of field robotics and review some of the limitations of the data.

Apart from simply benchmarking decision-making, we hope that this dataset can be further useful to researchers seeking to answer critical questions around using signage for mapless navigation in novel environments, including:
\begin{description}
\item[Q1.] How can navigational signage be perceived and interpreted, and how can its instructions be grounded to concrete actions?
\item[Q2.] How can common-sense priors be useful for interpreting navigational signage and grounding their instructions?
\item[Q3.] How can navigational systems effectively identify and recover from failures in decision-making? 
\item[Q4.] How can complex, hierarchical human goal specifications be related to the observed navigational cues? 
\end{description}

% Through this dataset, we hope that researchers can address critical questions about mapless, sign-centric navigation:
% \textcolor{red}{1.perception, how to ground
% 2.goal specification how to decompose complex instruction
% 3.common-sense - matching goal to sign locations 
% 4.failure recovery and what memory we need}
%     \paragraph{} \textbf{Q1:} What is the spatial-symbolic representation required for robust visual sign grounding?
%     \paragraph{}  \textbf{Q2:} How should we represent long-term spatio-temporal memory to address the complexity and hierarchical nature of human language goal specification?
%      \paragraph{}  \textbf{Q3:} What is the importance of semantic reasoning over complex and hierarchical human language goal specification? What is the importance of semantic reasoning over navigational signs?
%    \paragraph{} \textbf{Q4:} What is the significance of failure recovery for long-range mapless navigation?

More information is available on the dataset's website \footnote{\url{https://sites.google.com/view/signgapore-dataset?usp=sharing}}.

\begin{figure*}
\centering
    \begin{subfigure}{\textwidth}
        \centering
        \includegraphics[width=0.99\textwidth]{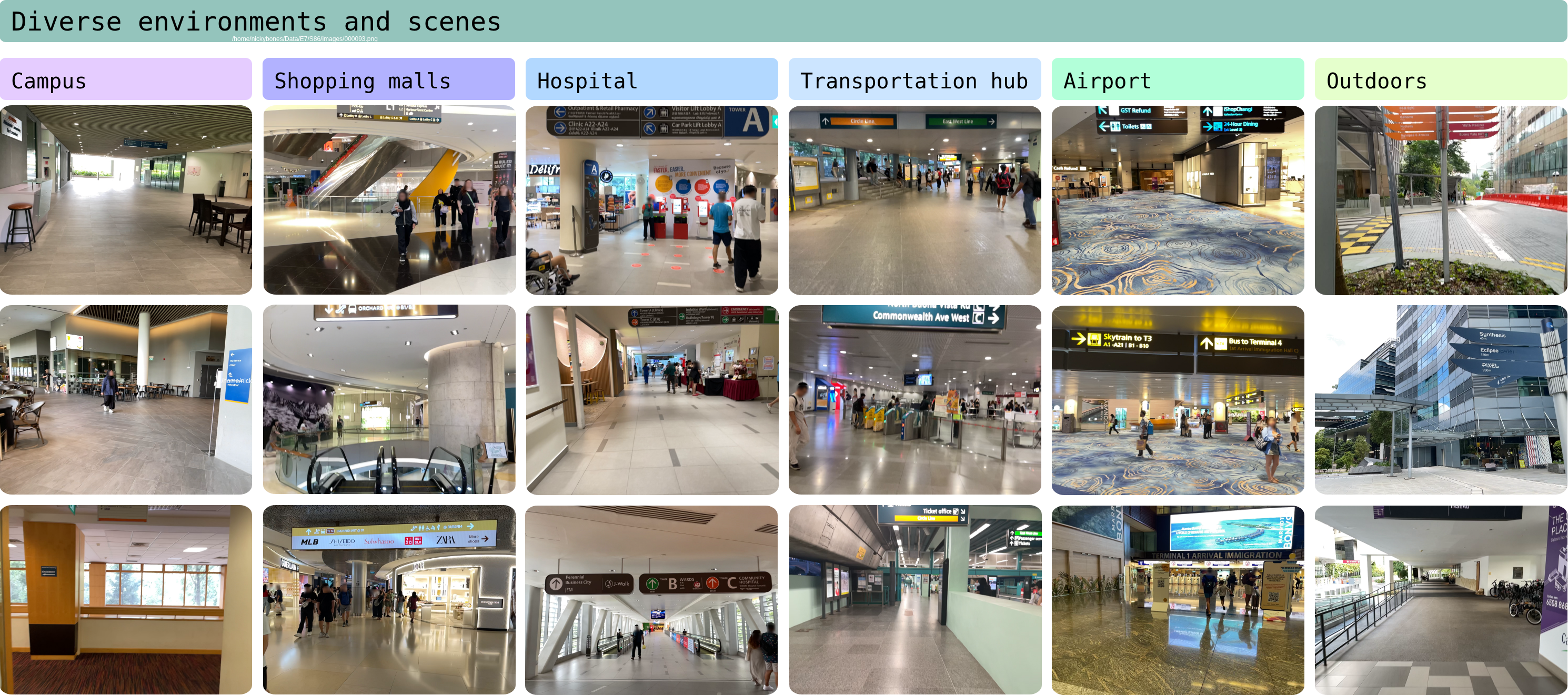}
        %{pics/SceneDivresity.png}
       \caption{Our dataset spans various public spaces that are frequented by humans. The dataset ranges from narrow indoor corridors, to vast open-space atriums and urban outdoor scenes. We captured complex scenes with multiple paths which provide a challenge in grounding directions to real-world structures. We also record scenarios with varying levels of dynamic activity, such as moving pedestrians, and distinct visual appearances, such as highly textured carpets and reflective and transparent surfaces. The scenes often include distractors, such as advertisements, display screens and non-navigational signs.       }
       \label{fig:scenes}
    \end{subfigure}
    
    \vspace{5mm} % Vertical gap between subfigures
    
    \begin{subfigure}{\textwidth}
        \centering
       \includegraphics[width=0.99\textwidth]{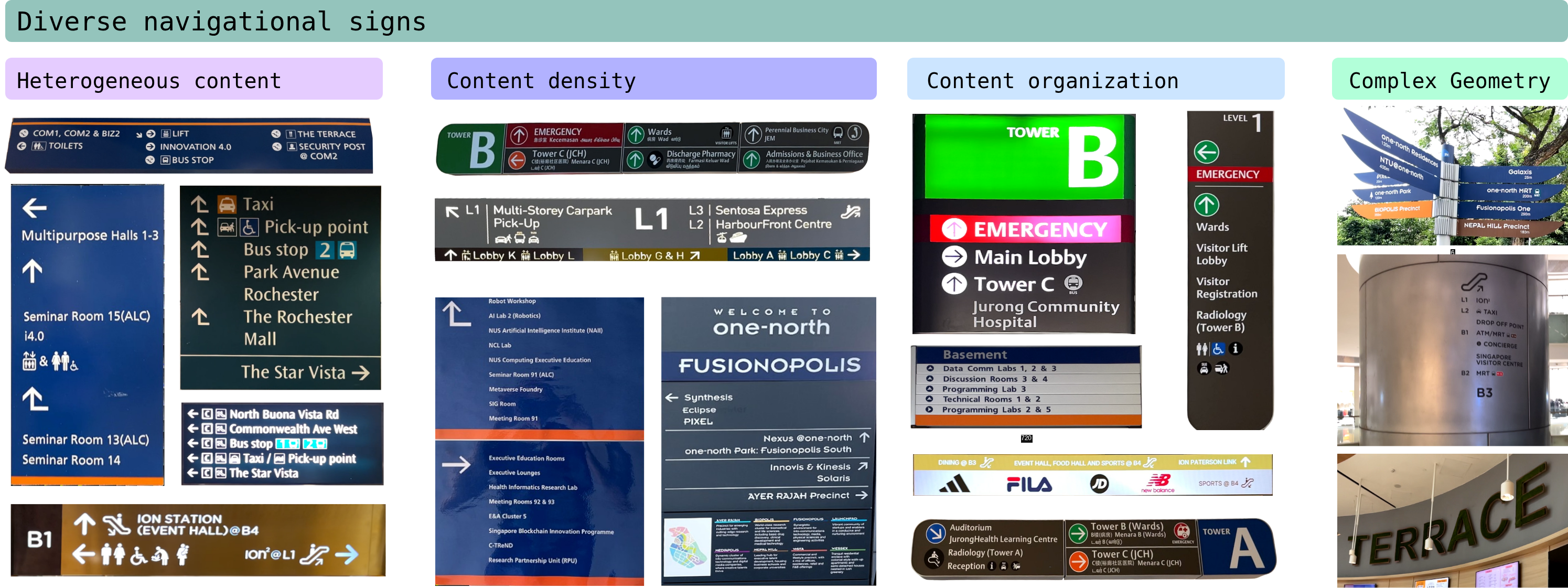}
       %{pics/signdiversity.png}
       \caption{Sign diversity is expressed in different aspects. Our dataset includes heterogeneous content, specifying locations with both text and symbols, as well as complex navigational cues that specify macro-actions by composing multiple symbols. We captured signs with high content density, including multilingual locations, hierarchical location information such as floor numbers and distracting elements. We also showcase various ways in which information can be encoded in signs, with different designs and association patterns between locations and navigational instructions. The dataset also features signs with non-planar geometry.        }
       \label{fig:signs}
        \end{subfigure}
\caption{\textbf{Diversity of signs and scenes.} Our dataset has diverse mission scenarios spanning a range of complex signs and scenes. }

% \textbf{(a)} contains multiple navigational cues, with distractors in the background. \textbf{(b)} has complex navigational cues that specify macro-actions by composing multiple symbols. Joint reasoning over other cues and the scene is needed in \textbf{(c)}, in order to disambiguate and ground cues such as that for the ``Auditorium''. \textbf{(d)-(e)} are complex scenes with multiple paths which provide a challenge in grounding directions to real-world structures. \textbf{(d)} also include 3D signage, where a sign is constructed from multiple non-parallel planes. }
\label{fig:diversity}
\end{figure*}

\begin{figure}
\centering
\includegraphics[width=0.99\columnwidth]{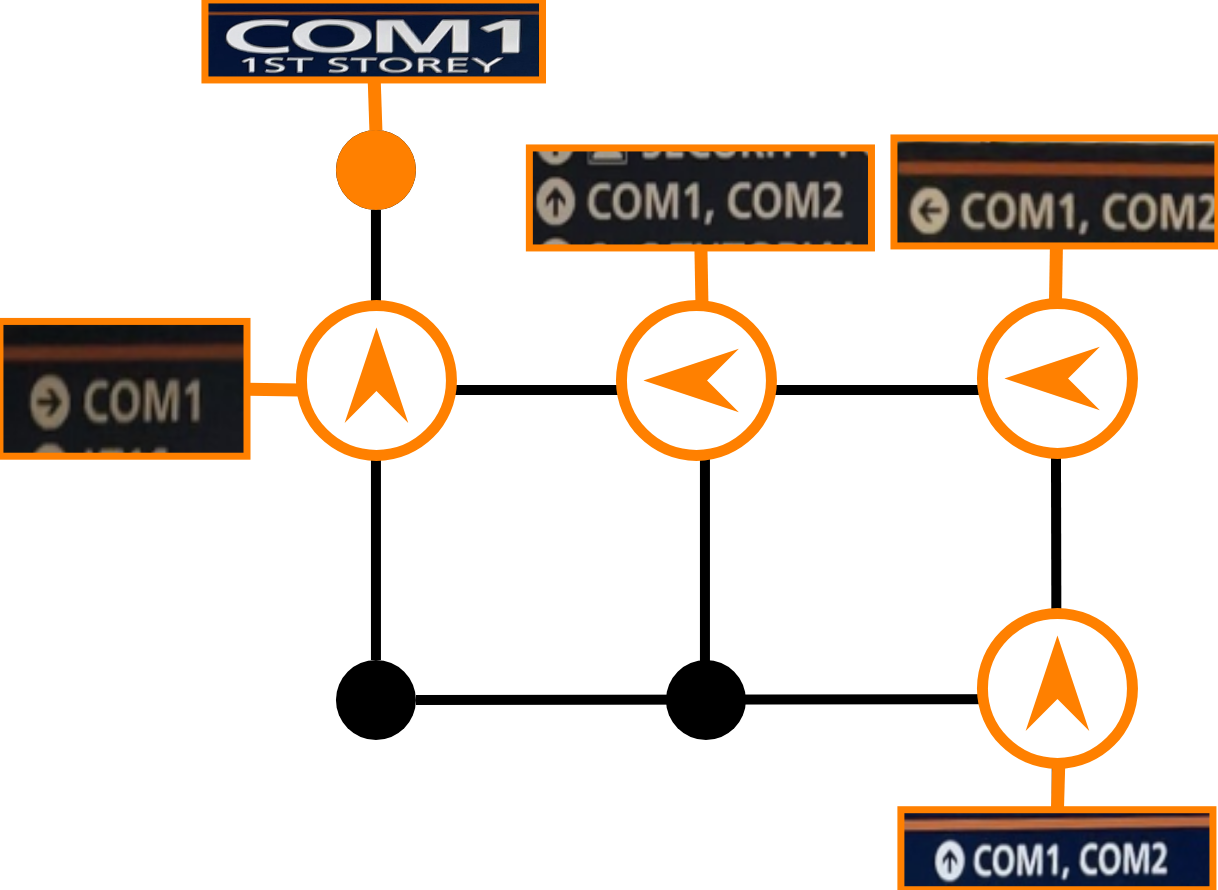} 
%{pics/SignOptimal.png}
\caption{\textbf{Sign-optimal path.} Given the mission goal ``COM1'', the robot must follow the sign-optimal path, which agrees with the navigational instruction associated with ``COM1'' at every step of the navigation process. Other paths can also lead to the goal, but they are not considered sign-optimal, despite having equal (or lesser) step length. }
\label{fig:signoptimal}
\end{figure}

\section{Related Work}
\subsection{Sign-based Navigation}
Mapless navigation is an increasingly important ability for autonomous agents~\citep{anderson2018cvpr,ku2020emnlp,zhang2025rss}. Navigational signs are powerful navigational cues embedded in human-oriented environments to guide wayfinding~\citep{gibson2009pap, richter2017springer}. However, use of these cues for robotic navigation remains underexplored~\citep{agrawal2026icraws, sun2026arxiv}. 

Recently, several works have begun to tackle aspects of this challenge. \cite{cao2026arxiv} proposed a sign-guided navigation approach, but assume that the semantic content of the signs is already extracted. Similarly, \cite{liang2020iros} and \cite{chandaka2025corl} investigate sign-based navigation in simplified settings, where the perceptual challenge of interpreting signs is largely circumvented. They make simplifying assumptions about the relative positioning of destinations and instructions on signs. 
They also only considered three canonical directions ($\leftarrow, \uparrow, \rightarrow$), without considering other types of directions or other symbol-based instructions (e.g. up-the-escalator, as seen in \figref{fig:signs}). 
As \cite{agrawal2026icraws} observed, these assumptions often do not hold true for complex, real-world signage---signs often capture information using non-standard symbols, have diverse styles and presentation, and may require reasoning and common-sense to interpret and associate destinations with instructions. 
They formalized the problem of \textit{Navigational Sign Understanding}, which aims to parse general navigational signs into destination–instruction pairs. Using these parsed pairs, \cite{zimmerman2026ral} globally localized in large-scale topological maps, by matching the spatio-semantic information extracted in these pairs to nodes in the graph. \cite{zimmerman2026isrr} further extended the \textit{Navigational Sign Understanding} problem beyond cardinal directions, to include navigational instructions such as ``up-escalator''. They also formulated the problem of \textit{Visual Sign Grounding}, which maps a location and its associated navigational instruction into a region in the sign's local environment. 

Prior work motivates sign-based navigation as a vital capability, and acknowledges the difficulty of interpreting and grounding signs from visual input. In particular, few approaches have yet evaluated their performance over long-horizon, sequential decision-making tasks. We propose \dataset{} to address these key issues.

\subsection{Simulated Visual Navigation Benchmarks}

Existing embodied navigation benchmarks provide controlled and reproducible environments for evaluating visual navigation capabilities. \cite{sun2026arxiv} extend the Habitat simulator~\citep{savva2019iccv} with real-world scans from public spaces. The simulation is intended for testing behavioral controllers and highlights signs as an important source of information for decision making. However, the environments conform to a Manhattan world structure and the behavior primitives are limited to $(\leftarrow, \uparrow, \rightarrow)$, reflecting the simplistic, visually uniform and auto-generated signage that only includes these 3 directions and no goal locations or symbols. In essence, the arrows were used as a visual stimuli for behaviors, without associating a goal with the desired navigational instruction. \cite{lin2025arxiv} provides more realistic embodied simulation with Isaac Sim~\citep{nvidia2026software}. Both simulation benchmarks are limited to static, generic and limited environments that do not reflect the difficulty of navigating in crowded, visually and topologically distinct human-oriented spaces and does not include realistic signage.

\subsection{Real-world Visual Navigation Benchmarks}

Instead of procedurally generating scenes or reproducing real-world scenes in simulation with some loss in fidelity, Visual Question Answering (VQA)-style datasets aim to evaluate decision-making approaches directly on real-world data from complex environments. NaviTrace~\citep{windecker2025arxiv} is a recent VQA-style benchmark for evaluating whether VLMs understand embodied navigation: given an image, a natural-language instruction, and an embodiment type such as human, legged robot, wheeled robot, or bicycle, the model must output a 2D navigation trace, with results showing that current VLMs still lag behind humans due to poor spatial grounding and goal localization. Nevertheless, NaviTrace evaluates each scenario in isolation, and it does not include a long-range navigation mission that requires sequential decision-making.

\section{Benchmark Tasks and Intended Use}
Navigational signs can enable navigation in unfamiliar places, when no prior maps are available. Therefore, we believe that a robot's ability to understand and ground navigational signs relates to its mapless navigation capabilities. For this reason, we offer this dataset for evaluating sign-based mapless navigation. The dataset is intended to evaluate three tasks: (i) sign understanding~\cite{agrawal2026icraws}; (ii) visual sign grounding~\cite{zimmerman2026isrr}; (iii) multi-step decision making for navigation.

This dataset is not intended for model training; it is intended as a held-out real-world evaluation benchmark. Unlike training-oriented navigation datasets that prioritize raw trajectory volume, \dataset{} is designed as a curated evaluation benchmark. Its unit of annotation is the sign-centric decision scenario: each scenario is manually grounded, connected within a topological transition graph, and incorporated into long-horizon missions.

% For clarity, we define navigational sign as a navigational aid that contains information about places and their associated navigational instruction. This spatio-symbolic information can be extracted as a collection of tuples, where each tuple associates a location $L_j$ with a navigational instruction $I_j$ .
% \textcolor{red}{define sign}

\subsection{Navigational Sign Understanding}
Following the definition of \textit{Navigational Sign Understanding} from \cite{agrawal2026icraws}, we define a navigational sign as a navigational aid that contains information about places and their associated navigational instruction. This spatio-symbolic information can be extracted as a collection of tuples, where each tuple associates a location $L_j$ with a navigational instruction $I_j$ . For the benchmark evaluation, we compare the location-instruction associations of selected locations with manually labeled ground truth labels.

\subsection{Visual Sign Grounding}
We borrow the formulation for \textit{visual sign grounding} from \cite{zimmerman2026isrr} and define it as a mapping of a location and its associated navigational instruction to a region in the sign's local coordinate system. For the benchmark evaluation,  we frame the visual sign grounding problem as a multiple-choice question where path options are provided~(\figref{fig:vqa}) for a queried location. 
The format of the VQA and how it can be processed is detailed in \secref{sec:description}.

\subsection{Multi-Step Decision Making for Navigation}

While \cite{zimmerman2026isrr} viewed the visual sign grounding problem for each scenario in isolation, this dataset introduces a multi-step VQA challenge, chaining decision making and action prediction into a long-range visual navigation task. The action prediction in each scenario will dictate which scenario will be encountered next, with the purpose of completing multi-step sign-based navigation to a destination. We refer to multi-step VQAs driven by a destination as "missions", and they are the primary focus of this dataset.

\section{The \dataset{} Dataset}

This section provides an overview of the environments in
which the dataset was collected, and describes our hand-held setup  and defines the metrics for evaluation. We also provide details on the quality of annotations used to evaluate the benchmark.

\subsection{Environments}
To ensure the collection of a diverse, representative dataset, we record data in multiple public spaces in Singapore, including malls, campus buildings, hospitals and transportation hubs, as well as outdoor environments. These environments are representative of places humans frequent during their daily activities, and we believe that accurately reasoning about signs in such environments open the door to city-scale autonomy. 

% \begin{figure*}
% \centering
% \includegraphics[width=\textwidth]{pics/sign_variety.jpg}
% \caption{\textbf{Diversity of signs and scenes.} Our dataset has diverse mission nodes spanning a range of complex signs and scenes. \textbf{(a)} contains multiple navigational cues, with distractors in the background. \textbf{(b)} has complex navigational cues that specify macro-actions by composing multiple symbols. Joint reasoning over other cues and the scene is needed in \textbf{(c)}, in order to disambiguate and ground cues such as that for the ``Auditorium''. \textbf{(d)-(e)} are complex scenes with multiple paths which provide a challenge in grounding directions to real-world structures. \textbf{(d)} also include 3D signage, where a sign is constructed from multiple non-parallel planes. }
% \label{fig:sign_variety}
% \end{figure*}

We ensure that our collected ``missions'' capture a diversity of decision-making points and scenarios. As illustrated in \autoref{fig:diversity}, we focus on two dimensions: sign diversity and scene diversity. First, capturing navigational signs with diverse visual appearances, complex layout, containing both textual and symbolic location description, and covering a wide range of navigational instructions, from the eight cardinal directions, to more symbolic instructions such as ``up-escalator'' and even compound instructions for multi-step actions. Second, we enrich scene diversity by collecting data in visually and topologically distinct environments. The recorded scenes represent a variety of topologies, far beyond the generic T-junction or X-junction configuration connected by corridors. The scenes include open spaces, curving paths and vertical connectors such as stairs, escalators and lifts. Additionally, the appearance varies between locations, including challenging features such as glass walls and railing, floors with varying textures, semantically defined paths such as paved roads surrounded by grass. Lastly, crowd density is also scene-dependent. 
We also ensure that the missions represent distinct challenges for long-horizon navigation, such as navigation between indoor and outdoors environments, interconnected buildings and across multiple floors, as detailed in \tabref{tab:environments}. Visualization of selected mission trajectories is presented in \figref{fig:mission}, and a more detailed illustration of a mission's sign-optimal path is provided in \figref{fig:signoptimal}.

\subsection{Data Collection Platform}
To make the dataset as general as possible, and suitable for diverse platforms, we chose a minimal sensor setup, which includes sparse depth, RGB, odometry and IMU measurements. As the dataset is collected in crowded public spaces, the data collection process should minimize disruption and discomfort, leading us to choose an iPhone 13 Pro.  
The dataset was collected using the StrayScanner~\citep{}, a free open-source software for iOS for recording of raw LiDAR, RGB, odometry and IMU data. StrayScanner is built on top of ARKit~\citep{applearkit2026software}, Apple's augmented reality framework. ARKit was used to collect multiple RGB-D datasets such as ARKitScenes~\citep{baruch2021nips} and ARKitTrack~\citep{zhao2023cvpr}, and is known for its highly accurate low-drift VIO~\citep{kim2022sensors}. 
StrayScanner allows the user to choose a frame-rate  between 1 and 60 FPS, and utilizes ARFrames to synchronize depth, RGB and VIO odometry measurements at the selected FPS, with IMU measurements logged at 100~\unit{\hertz}. The depth frames captured by the Time-of-Flight sensor are aligned with the RGB frames, but with a lower resolution. The frame transformation between the IMU and the RGB sensor is handled by Apple's API, and the IMU data is logged in the unified camera tracking frame. Therefore, no coordinate frame transformations are required when processing the dataset. 
Overall, more than a quarter of a million frames were collected in this dataset, equivalent to 5 hours of footage from sign-centric scenarios in various environments. More details about the raw data can be found in \tabref{tab:rawdata}. To address privacy concerns, the dataset was anonymized by blurring faces.

\begin{figure}[htbp]
    \centering
    % First Subfigure
    \begin{subfigure}[b]{0.32\columnwidth}
        \centering
        \includegraphics[width=\columnwidth]{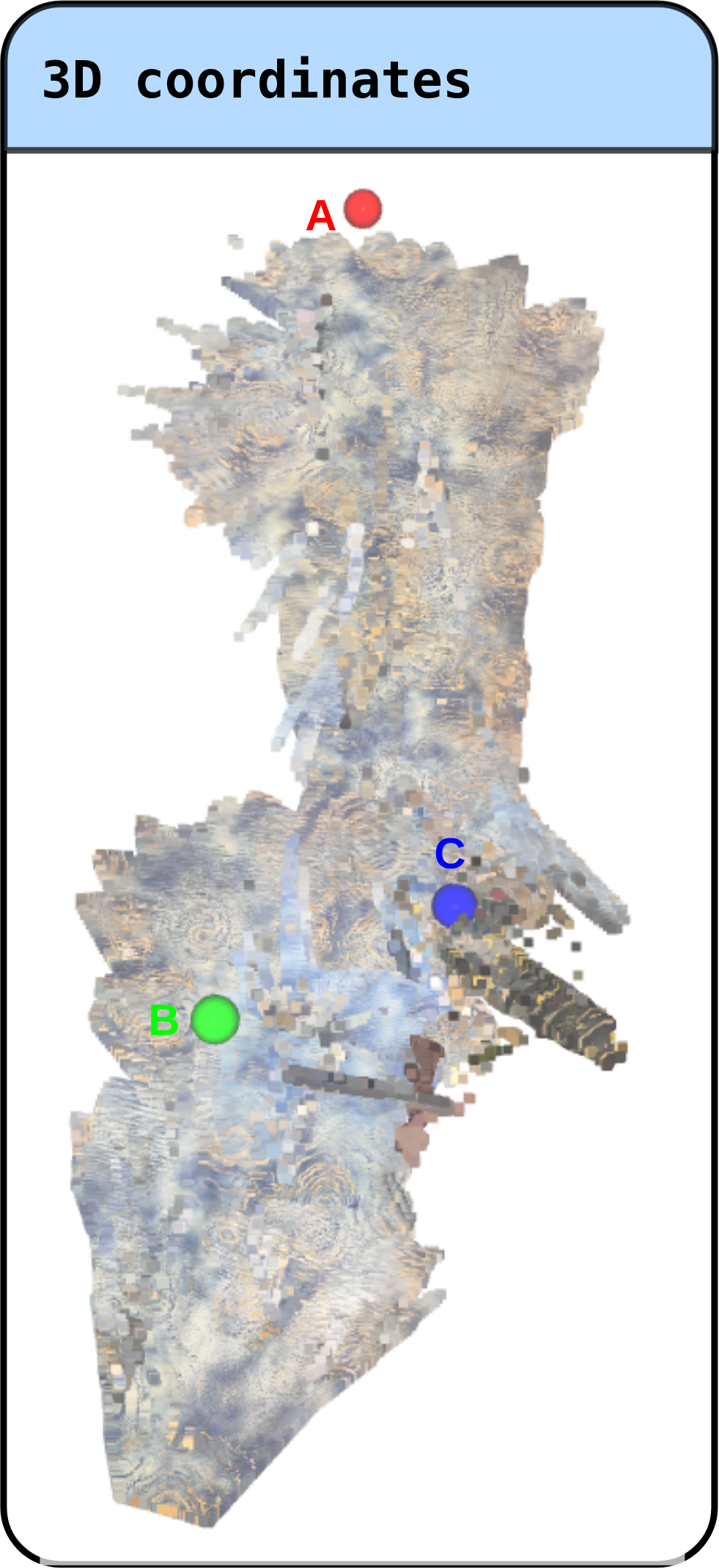}
        %{pics/VQA3D.png}
        \captionsetup{justification=centering} 
        %\caption{VQA options as 3D points in the robot's coordinate frame.}
        \caption{}
        \label{fig:vqa3d}
    \end{subfigure}
    \hfill % Adds horizontal spacing between the two images
    % Second Subfigure
    \begin{subfigure}[b]{0.3\columnwidth}
        \centering
        \includegraphics[width=\columnwidth]{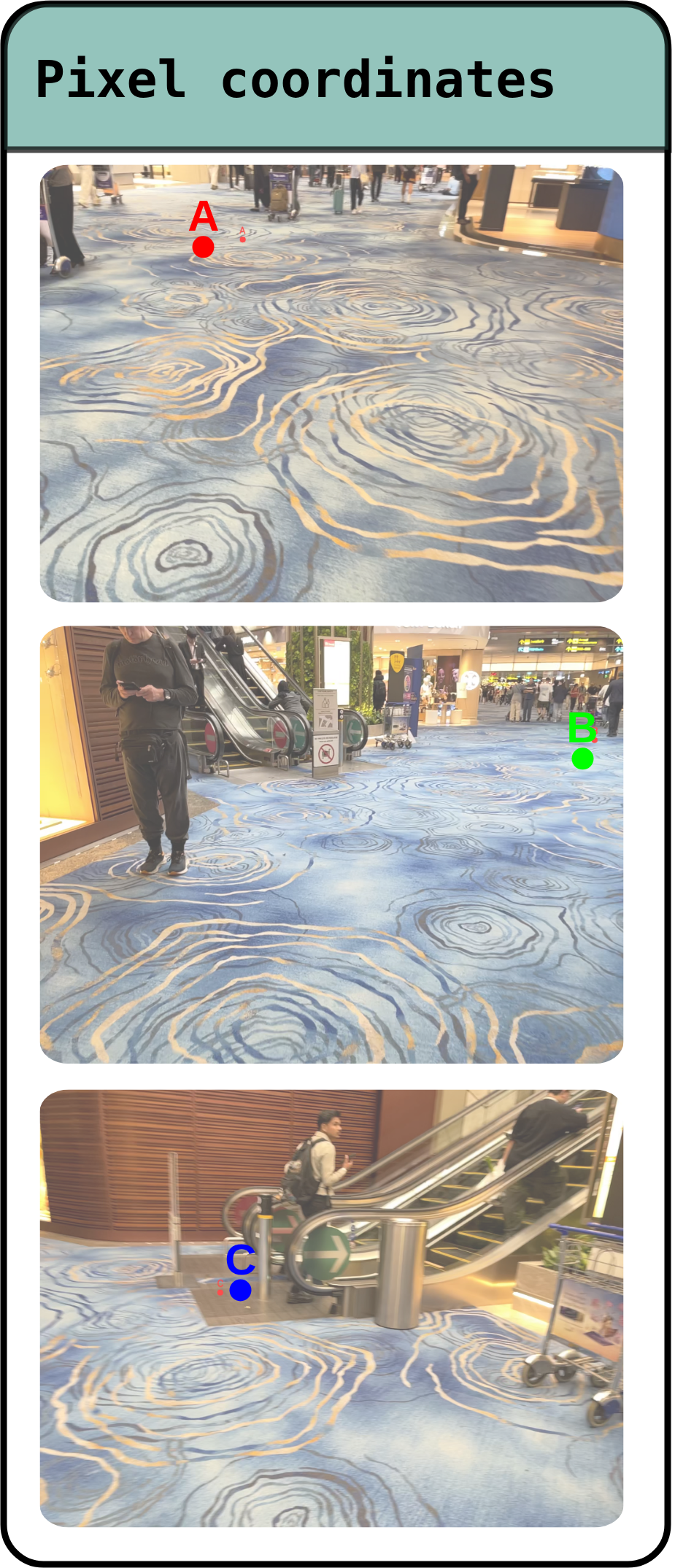}
        %{pics/VQA2D.png}
       % \caption{VQA options in pixel coordinates across multiple, different view frames.}
       \captionsetup{justification=centering} 
         \caption{}
        \label{fig:vqa2d}
    \end{subfigure}
\hfill
     \begin{subfigure}[b]{0.346\columnwidth}
        \centering
        \includegraphics[width=\columnwidth]{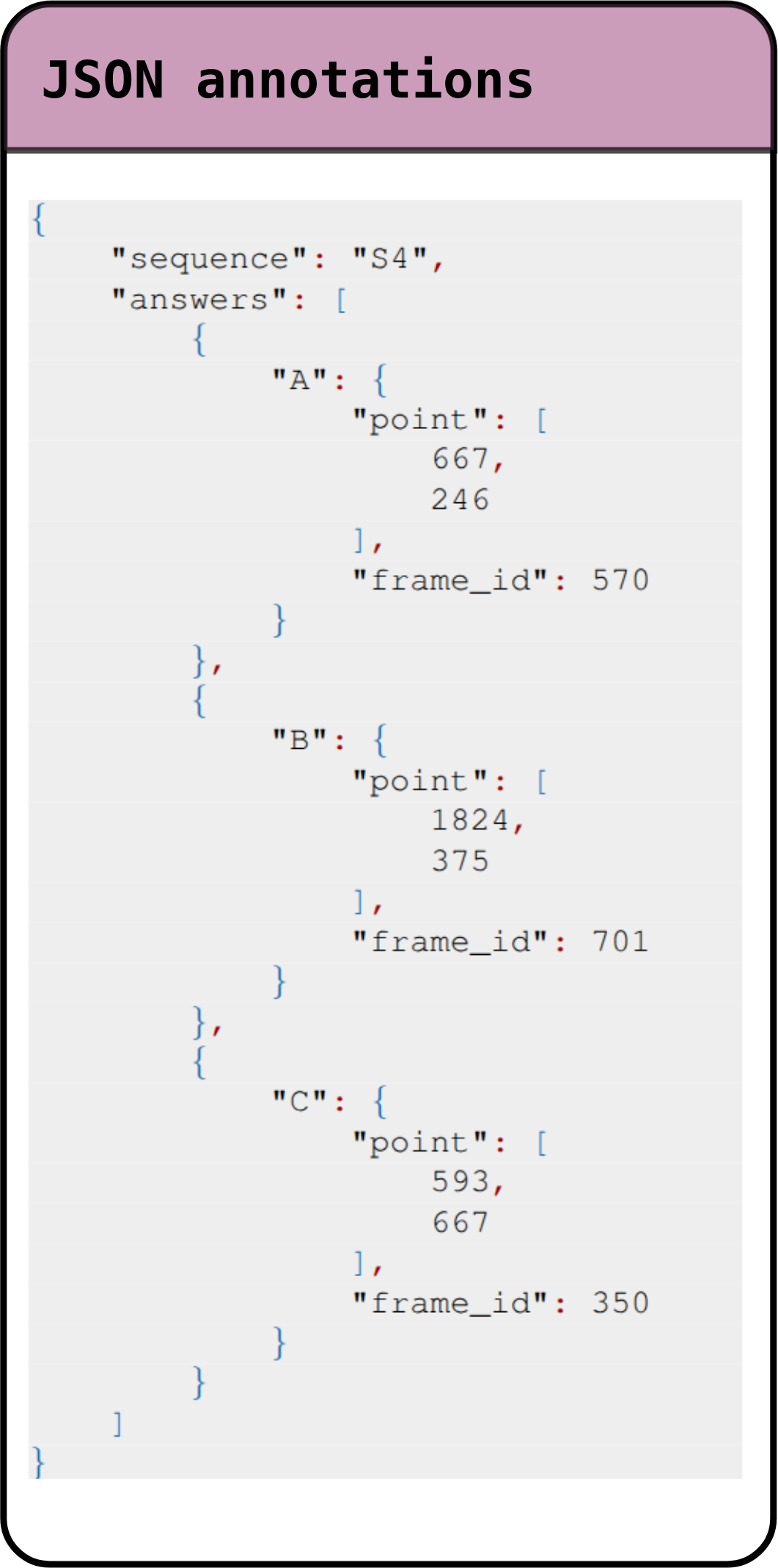}
        %\caption{VQA op
        %{pics/VQAJson.png}
        %\caption{VQA options in a json format.}
        \captionsetup{justification=centering} 
         \caption{}
        \label{fig:vqajson}
    \end{subfigure}
    
    % Main Figure Caption
    \caption{\textbf{VQA format}. The VQA options are provided in 3 different formats, to accommodate a variety of methods, from modular to end-to-end approaches: (a) as 3D points in the robot's coordinate frame; (b) in pixel coordinates across multiple different view frames; (c) in a JSON format. }
    \label{fig:vqa}
\end{figure}

% \begin{figure}[htbp]
%     \centering
% \includegraphics[width=0.99\columnwidth]{pics/vqaformat.png}
%     % Main Figure Caption
%     \caption{\textbf{VQA format}. The VQA options are provided in 3 different formats, so accommodate a variety of approaches, from modular to end-to-ends.}
%     \label{fig:vqa}
% \end{figure}

% \begin{figure}
% \centering
% \includegraphics[width=0.99\columnwidth]{pics/TopoGraph2.pdf}
% \caption{\textbf{Sign-centric topological graph.} Each environment is a topological graph, where each node is a scenario that contains measurements of the scene and observations of the sign. The edges represent possible actions afforded by the VQA options. }
% \label{fig:topograph}
% \end{figure}

\subsection{Metrics}
%\textcolor{red}{explain to what scenarios we provide GT}
To evaluate the performance on the benchmark, a few metrics are considered.  
\paragraph{Success Rate}
For the multi-step VQA, the goal is to reach the destination. Therefore, success rate measures the completion of missions, and it does not restrict the solution to the sign-based optimal path.
\paragraph{Strict Success Rate}
In addition to completing this mission, this metric considers a success to be a solution identical to the sign-based optimal path~(\figref{fig:signoptimal}). If there were grounding errors along the execution, but the mission was completed through failure recovery, it would still be considered a failure.
% \paragraph{Trajectory Length}
% Successful grounding for each VQA will result an optimal path to the destination, but mission can be completed even with imperfect grounding, given failure handling. Therefore, we also report the trajectory length, and score it with respect to the optimal path length.
\paragraph{VQA Accuracy}
We evaluate the accuracy of single-step visual sign grounding, and provide grounding ground truth for all scenarios that are included in at least one of the missions. 
\paragraph{Sign Parsing Accuracy}
We evaluate the sign parsing accuracy for destinations that are included in our missions, and provide ground truth parsing to all scenarios whose sign contains a mission goal location. 3D signage~(\figref{fig:signs}) is excluded from the labels as it cannot be labeled in 2D.
% Some missions require higher levels of reasoning, when signs do not include that exact destination name (i.e. for destination "LT18", some signs will include "LT16-LT19"). 

\subsection{Post-Processing}

Each environment in the dataset is represented as a sign-centric topological graph~(\figref{fig:scenariodyn}), where edges represent navigational behaviors, such as turning right, going up the stairs, etc. Nodes in the graph belong to one of two types; scenario nodes that contain a sequence recorded around a sign, and terminal nodes representing a destination point.  Each scenario node includes raw data and a multiple-choice VQA-style annotations indicating the potential actions. Once an action is chosen, the next scenario is rolled out based on the sign-centric topological graph. RGB frames were anonymized by blurring faces.

The benchmark is primarily evaluated on missions - multi-step VQAs driven by a destination. For each mission with a starting node and destination, there exists an optimal path on the graph to the goal node, based on the navigational instructions encoded in the signs. Upon reaching a terminal node that corresponds to the destination, the mission is completed. 
Both the sign-centric topological graph and the VQA are manually annotated. 

For each environment we attempt to collect as many signs as possible and achieve high coverage. From the places indicated in the signs, we sample our destinations. We create a subgraph with all nodes in the topological graph that contain a reference to a specific destination. We then select the furthest node as the starting node and compute the sign-optimal path~(\figref{fig:signoptimal}) on the subgraph as our ground truth for the multi-task VQA.

Our missions can be categorized into three different levels in terms of reasoning effort. The easiest missions include a simple goal description such as ``COM1'', which appears explicitly in all nodes on the optimal path. The medium-effort missions include a simple goal description such as ``COM2'', which may appear implicitly in signs under descriptions such as ``COM1-COM3''. The hardest missions include a complex language instruction such as ``go to programming lab 4 in the COM1 building''. This requires decomposition to subgoals prior to the visual sign grounding.

The manual and physical collection and annotation efforts are necessary because the target phenomenon is the structure of real human navigation infrastructure, not a procedurally generated approximation of it. Overall, 456 scenarios were annotated and 56 missions were designed. 

\subsection{Annotations and Quality Control}
All benchmark annotations were created manually. For each scenario, a human annotator reviewed the complete recorded sequence together with the navigational sign and the set of possible navigation goals. The annotator then selected a representative RGB frame and a pixel coordinate for each visual grounding option. Candidate options were chosen to be visually and spatially distinct and to correspond to plausible groundings of a navigational instruction in the local scene. To ensure that the benchmark also tests the ability to reject incorrect interpretations, some candidates were deliberately placed along paths that lead to dead ends or otherwise do not provide a valid continuation toward the queried destination.

The correct grounding for each query was selected manually by a human annotator. We exclude scenarios for which either the interpretation of the sign or its grounding in the surrounding scene is ambiguous. Consequently, each annotated query has a single intended ground-truth grounding among the provided candidates. The annotations were subsequently reviewed by additional human annotators and further checked through the benchmark evaluation pipeline to identify inconsistencies between grounding annotations and graph transitions.

The sign-centric topological graphs and mission annotations were also manually verified. Connectivity was checked by simulating transitions between scenario nodes according to the annotated actions and confirming that the resulting transitions agree with the physical connectivity of the environment. For each mission, a human annotator specified the ground-truth sign-optimal path~(\figref{fig:signoptimal}) by following the navigational instructions encoded in the observed signs. The computed sign-optimal paths were then compared against these manually specified paths to verify both the graph connectivity and mission annotations.

\begin{table*}[h]
\small\sf\centering
\caption{\textbf{Raw data.} Summary of the raw data collected in this dataset.\label{tab:rawdata}}
\begin{tabular}{llll}
\toprule
Sensor& Data Type & Data logged & Frequency\\
\midrule
iPhone Camera & 1920 $\times$ 1440 RGB images & 274742 & 15~\unit{\hertz}\\
iPhone Lidar & 256 $\times$ 192 depth images & 274742 & 15~\unit{\hertz}\\
iPhone IMU & Tri-axial linear accelerations, tri-axial
angular velocities & 1772208 sets & 100~\unit{\hertz}\\
iPhone Odometry &  (x, y, z, q\textsubscript{x}, q\textsubscript{y}, q\textsubscript{z}, q\textsubscript{w}) & 274742 & 15~\unit{\hertz}\\
\bottomrule
\end{tabular}
%\\[10pt]
\end{table*}

\section{Using the \dataset{} Dataset} \label{sec:description}
% How the reader can parse (programmatically) and index into the data
% Where appropriate, ground truth data should be provided and details given on how it was obtained and how accurate it is.
% Where possible, numerical data should be provided in human readable plain text format, images in high quality jpeg or png format. Data papers based on solely on rosbags and other changing binary formats will be returned without review (these binary formats can be provided as supplementary material if desired)

This section summarizes the contents of the dataset and
describes the utility scripts and tools we have provided to
process the data. Our code is available on Github \footnote{\url{https://github.com/FullMetalNicky/SiGNgapore/tree/main}}.

As indicated by \tabref{tab:environments}, the dataset was collected in a total of 7 environments. Each environment contained multiple missions~(\tabref{tab:environments}), each mission composed of multiple goal-oriented VQA scenarios. 
All of the data logged were converted into human-readable formats.
In the dataset archive each environment has its own directory, $\{E_i\}_{i=1}^7$, which contains all the scenarios $\{S_i\}_{i=1}^n$ recorded in this environment. 
An environment directory also contains the directories , $\{T_i\}_{i=1}^k$, each with its own \texttt{mission.json} file. 

\paragraph{Missions:} Each mission is described in a  \texttt{mission.json} file, which is characterized by the destination name and its associated terminal node, the start node and the action-based state transition matrix. The agent starts the mission from a scenario node, and based on the action selected in VQA, progresses to the next node conditioned on the state transition matrix. 

\paragraph{Scenarios:} Each scenario directory shares the same structure as described in \tabref{tab:format}. For \texttt{imu.csv} and \texttt{odometry.csv}, each comma-separated values (CSV)
file contains a header line describing the fields in the file, including a time-stamp for every measurement. A frame ID is associated with each timestamped odometry measurement in \texttt{odometry.csv}, and this same index is used to name the associated RGB, depth and confidence maps. 
The depth, RGB and VIO data is synchronized using the ARKit framework. We provide scripts(~\secref{sec:scripts}) to associate the RGB and depth frames with their time-stamp. For each sequence file, the camera intrinsics are provided by the
device manufacturer and are included in \texttt{camera\_matrix.csv}. %\textcolor{red}{Prior to data logging, the accuracy of these parameters was verified using ?}.
Due to storage considerations, the RGB images are stored as a HEVC encoded video, rgb.mp4, but we provide a script~(\secref{sec:scripts}) to extract the images to the format reported in ~\tabref{tab:format}. In addition to raw data, each sequence contains a VQA.
\paragraph{VQA} The VQA directory contains a \texttt{query.json} file~(\figref{fig:vqajson}) and a subdirectory named \texttt{samples/}. \texttt{query.json} encodes each option in the multi-choice VQA as a point in pixel coordinates in an indexed RGB frame. To cater to end-to-end approaches, these choices are also available as annotations~(\figref{fig:vqa2d}), and stored in the \texttt{samples/} directory. Using the utility scripts in \secref{sec:scripts}, the odometry and depth measurements can be recovered for each indexed frame, allowing the user to project these points to 3D, as visualized in \figref{fig:vqa3d}.

\subsection{Utility Scripts} \label{sec:scripts}
For storage considerations, the RGB image stream is stored as a video. We provide \texttt{Vid2Img.py} script to convert the video to an image folder according to the format described in \tabref{tab:format}. We provide two different ways to manipulate the logged raw data:
%For offline processing, we provide a Python class for data loading, and for online evaluation the data can be broadcast through a ROS2 node or converted to a ROS2 bag.
\begin{itemize}
    \item \textbf{DatasetLoader} - The class \texttt{DatasetLoader} provides a Python interface for loading the dataset. The class allows users to load all the data as synchronized and indexed Numpy array, that can be consumed frame by frame, as the odometry, RGB and depth images are already synchronized by ARKit through an ARKitFrame. %\textcolor{red}{something about IMU?}. 
    This script is suitable for manipulating the data in an offline manner.
    \item \textbf{Dataset2ROS2bag} - The script \texttt{Dataset2ROS2bag.py} allows users to convert a dataset sequence into a ROS 2 bag. Users can define their own topic names, and create a ROS 2 bag with RGB, depth, odometry and IMU measurements, as well as the TF tree. This format is suitable for processing the data in real-time.
\end{itemize}
% \subsubsection{DatasetLoader}
% The class \texttt{DataserLoader} provide a python interface for loading the dataset. The class allows users to load all the data as synchronized and indexed numpy array, that can be consumed frame by frame, as the odometry, RGB and depth images are already synchronized by ARKit through an ARKitFrame. \textcolor{red}{something about IMU?}. This script is suitable for manipulating the data in an offline manner.

% \subsubsection{Dataset2ROS2Bag} The script \texttt{Dataset2ROS2bag.py} allows users to convert a dataset sequence into a ROS2 bag. Users can define their own topic names, and create a ROS2 bag with RGB, depth, odometry and IMU measurements, as well as the TF tree. This format is suitable for processing the data in real-time.

\subsubsection{BenchmarkEngine}
We provide Python script for benchmark evaluation, \texttt{BenchmarkEngine.py}. This class allows the user to evaluate a single mission, a single environment or the entire dataset, and reports the performance according to the metrics.  
The benchmark engine advances the mission from scenario to scenario, based on the action selected by the evaluated approach. The benchmark requires an interface, which is derived from \texttt{MyAlgorithm} class, provided in the script. The arguments to the \texttt{ProcessScenario} function are the environment, mission and scenario IDs, and the return value is the selected option from the VQA. Based on the selected option, the benchmark engine will call \texttt{MyAlgorithm} with the next scenario.

\subsection{Venue Maps}
\begin{figure*}
\centering
\includegraphics[width=0.95\textwidth]{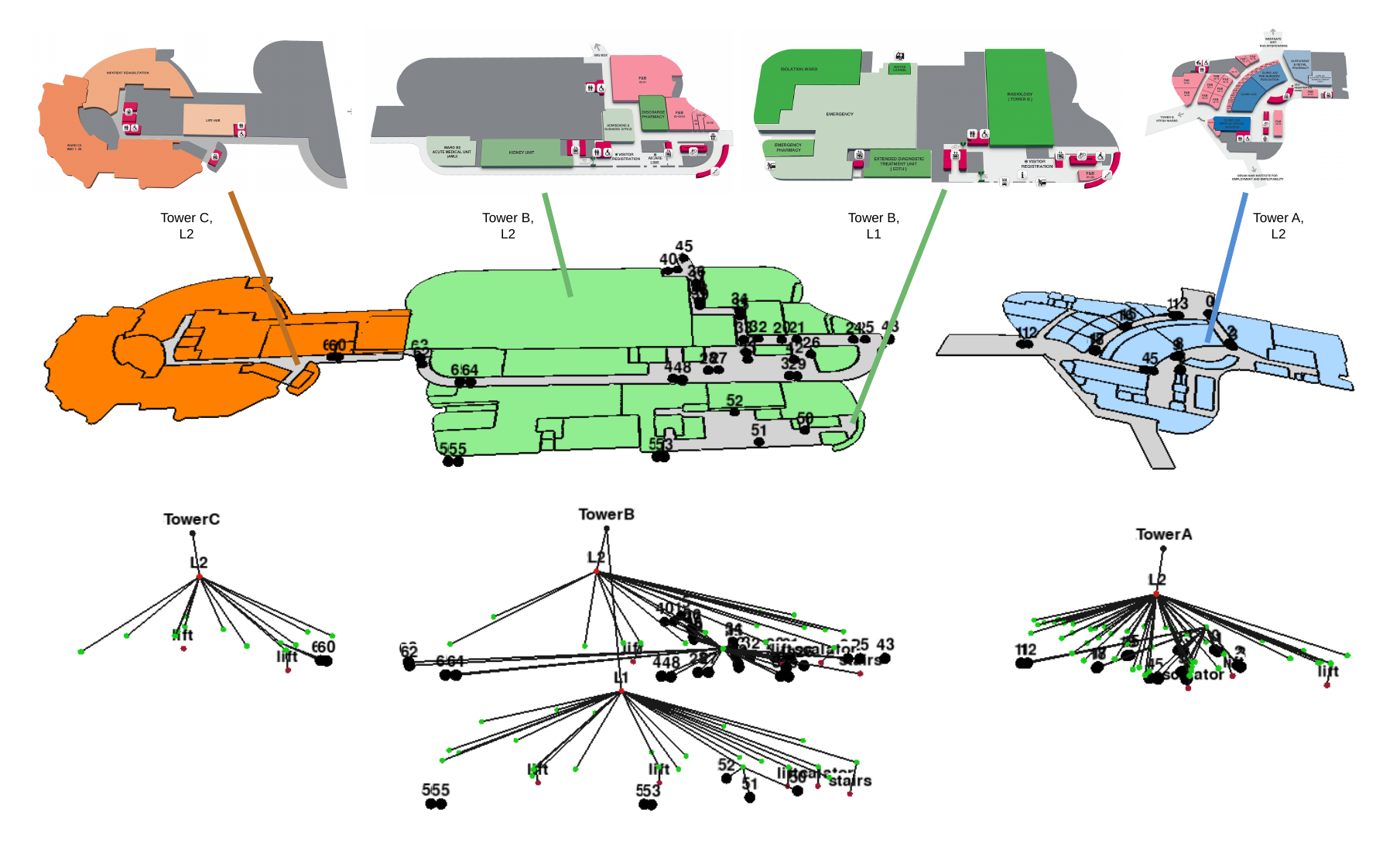}
\caption{\textbf{GPS-aligned SceneGraphs.} When available, we provide GPS-aligned SceneGrpah of the environment, annotated with building, floor and room nodes.}
\label{fig:hospital}
\end{figure*}

In addition to providing the raw data and the missions, we also provide venue maps when available.
These maps, combined with the raw data, can extend the utility of this dataset to other tasks, discussed in more detail in section \secref{sec:discussion}.
For our environments, the venue maps and floor plans, when available, are provided separately for each floor and each building, without alignment or proper scaling. In addition to providing the original maps, we also aligned and scaled the venue maps to GPS coordinates, such that they can be merged into one single environment. Based on the unified maps, we provide our own topo-metric representation, in the form of a SceneGraph, which includes building, floor and room nodes. For the map alignment and graph extraction, we rely on our previous work~(\cite{zimmerman2026ral}). The scenario locations are then annotated on these maps, available in pixel coordinates in each separate map file and GPS coordinates per environment. 
In \figref{fig:hospital}, we visualize the entire graph of the hospital complex, spanning three buildings and multiple floors. For environments where indoor maps are not available, or the scenarios were recorded outside, we provide approximate GPS coordinates for each scenario location, as well as visualizations on OpenStreetMap. 

\begin{table*}[h]
\small\sf\centering
\caption{\textbf{Environments and missions.} For each environment, we provide information about its area size, type, number of grounding scenarios and the missions. Each mission can span across different floors, interconnected building and urban outdoor spaces. The mission length refers to number of scenarios in the sign-based optimal path.\label{tab:environments}}
\begin{tabular}{c|c|c|c|c|c|c|c|c}
\toprule
Environment &  Scenarios & Area~(\unit{\meter \squared})  & Type & Missions & multi-building & multi-floor & Outdoor & Length \\
\midrule

\multirow{13}{*}{E1} & \multirow{13}{*}{74}  & \multirow{13}{*}{45,423} & \multirow{13}{*}{campus}  &  M1 & \xmark & \cmark  & \cmark& 4\\ 
& & & & M2 & \cmark  & \cmark & \xmark  & 8 \\ 
& & & & M3 & \cmark  & \cmark &\xmark &  6\\
& & & & M4 & \cmark  & \cmark  &\xmark & 9\\ 
& & & & M5 & \cmark  & \cmark  &\xmark & 7\\
& & & & M6 & \cmark  & \cmark  &\xmark & 7\\ 
& & & & M7 & \cmark  & \cmark   &\xmark &7\\
& & & & M8 & \cmark  & \cmark   &\xmark &6\\ 
& & & & M9 & \cmark  & \cmark  &\xmark & 8\\
& & & & M10 & \cmark  & \cmark &\xmark &  7\\
& & & & M11 & \cmark  & \cmark &\cmark &  6\\ 
& & & & M12 & \cmark  & \xmark  &\cmark & 5\\ 
& & & & M13 & \cmark  & \cmark  &\xmark & 16\\ \midrule
\multirow{4}{*}{E2} & \multirow{4}{*}{75}  & \multirow{4}{*}{142,854}  & \multirow{4}{*}{mall}  &  M1 & \cmark & \xmark & \xmark & 6\\ 
& & & & M2 & \cmark  & \xmark  &\xmark& 7 \\ 
& & & & M3 & \cmark  & \xmark &\xmark&   5\\
& & & & M4 & \cmark  & \xmark  &\cmark&  8\\  \midrule
\multirow{10}{*}{E3} & \multirow{10}{*}{58} & \multirow{10}{*}{37,088}  & \multirow{10}{*}{hospital}  &  M1 & \cmark & \cmark &\cmark &  9\\ 
& & & & M2 & \cmark  & \xmark &\xmark&  6 \\ 
& & & &M3 & \cmark  & \xmark &\xmark&   9\\
& & & &M4 & \cmark  & \xmark &\xmark&  9 \\ 
& & & &M5 & \cmark  & \xmark &\xmark&  7 \\ 
& & & &M6 & \cmark  & \xmark &\xmark&  8 \\
& & & &M7 & \cmark  & \xmark &\xmark&  11 \\ 
& & & &M8 & \cmark  & \xmark &\xmark&  8 \\
& & & &M9 & \cmark  & \xmark &\xmark&   13\\ 
& & & &M10 & \cmark  & \xmark &\xmark&   10\\ \midrule
\multirow{6}{*}{E4} & \multirow{6}{*}{61} & \multirow{6}{*}{109,542}  & \multirow{6}{*}{mall}  &  M1 & \xmark & \cmark &\xmark& 6 \\ 
& & & &M2 & \xmark  & \cmark &\xmark&  8 \\ 
& & & &M3 & \xmark  & \cmark &\xmark&  6 \\
& & & &M4 & \xmark  & \cmark &\xmark&  6 \\ 
& & & &M5 & \xmark  & \cmark &\xmark&  6 \\ 
& & & &M6 & \xmark  & \cmark  &\xmark& 6 \\ \midrule
\multirow{7}{*}{E5} & \multirow{7}{*}{45} & \multirow{7}{*}{426,139}  & \multirow{7}{*}{outdoors}  &  M1 & \xmark & \xmark &\cmark &  6 \\ 
& && & M2 & \xmark & \xmark& \cmark   &5\\ 
& && & M3 & \xmark & \xmark &  \cmark & 5\\
& && & M4 & \xmark &\xmark & \cmark   &8\\  
& & & &M5 &\xmark  & \xmark & \cmark  & 8\\ 
& & & &M6 & \xmark &\xmark & \cmark  &5\\
& & & &M7 & \xmark &\xmark & \cmark  & 9\\ \midrule
\multirow{7}{*}{E6} & \multirow{7}{*}{56} & \multirow{7}{*}{71,655}  & \multirow{7}{*}{transportation hub}  &  M1 & \cmark & \cmark &\cmark&  7\\ 
& & & &M2 & \cmark  & \cmark &\cmark&  7 \\ 
& & & &M3 & \cmark  & \cmark &\xmark&  6 \\
& & & &M4 & \cmark  & \cmark &\xmark&  6 \\ 
& & & &M5 & \cmark  & \cmark &\xmark&  7 \\ 
& & & &M6 & \cmark  & \cmark &\xmark&  8 \\ 
& & & &M7 & \cmark  & \cmark &\cmark&  9 \\ \midrule
\multirow{8}{*}{E7} & \multirow{8}{*}{87} & \multirow{8}{*}{308,000}  & \multirow{8}{*}{airport}  &  M1 & \xmark & \xmark &\xmark& 9 \\ 
& & & & M2 & \xmark  & \xmark &\xmark  & 13\\ 
& & & & M3 & \xmark  & \cmark &\xmark&  7\\
& & & & M4 & \xmark  & \xmark  &\xmark& 6\\ 
& & & & M5 & \xmark  & \xmark  &\xmark& 9\\
& & & & M6 & \xmark  & \xmark  &\xmark& 8\\ 
& & & & M7 & \xmark  & \xmark   &\xmark&7\\
& & & & M8 & \xmark  & \xmark   &\xmark&14\\
& & & & M9 & \xmark  & \cmark   &\xmark&7\\ %\midrule
%Total & 456 & meow & - & 56 & & & & \\
\bottomrule
\end{tabular}
%\\[10pt]
\end{table*}

% \begin{table*}[h]
% \small\sf\centering
% \caption{missions collected in this dataset.\label{tab:missions}}
% \begin{tabular}{lllll}
% \toprule
% Environment & mission & nodes & length & type \\
% \midrule
% \multirow{3}{*}{E1} & T1 & many & many & multi-building, multi-floor \\ 
% & T2 & many & many & multi-building, multi-floor \\ 
% & T3 & many & many & multi-building, multi-floor \\
% \midrule
% \multirow{3}{*}{E2} & T1 & many & many & multi-building, multi-floor \\
%  & T2 & many & many & multi-building, multi-floor \\
%  & T3 & many & many & multi-building, multi-floor \\
% \bottomrule
% \end{tabular}
% %\\[10pt]
% \end{table*}

\begin{table*}[h]
\small\sf\centering
\caption{\textbf{Scenarios.} Human-readable and image data formats and naming conventions for each scenario.\label{tab:format}}
\begin{tabular}{lll}
\toprule
File Name & Content  & Format \\
\midrule
images/000000-999999.png & 1920 $\times$ 1440 RGB image & PNG\\
depth/000000-999999.png & 256 $\times$ 192 depth image & PNG\\
confidence/000000-999999.png & 256 $\times$ 192 depth confidence map & PNG\\
imu.csv & \makecell[l]{Accelerometer and gyroscope \\
measurements} & \makecell[l]{Timestamp~(\unit{\second}), (x, y, z) linear accelerations~\unit{\meter \squared \per\second}, \\ (x, y, z) angular velocities~(\unit{\radian \per \second})} \\
odometry.csv & Local pose estimates & \makecell[l]{Timestamp~(\unit{\second}), frame id, (x, y, z) linear translation~\unit{\meter},\\ orientation (q\textsubscript{x} ,
q\textsubscript{y} , q\textsubscript{z} , q\textsubscript{w} )~\unit{\radian}} \\
camera\_matrix.csv & 3 $\times$ 3 camera intrinsics &  (f\textsubscript{x}, 0, c\textsubscript{x}, 0, f\textsubscript{y}, c\textsubscript{y}, 0, 0, 1) \\
% VQA/query.json & & \\
% VQA/samples/000000-999999\_ground.png & & \\
\bottomrule
\end{tabular}
%\\[10pt]
\end{table*}

\begin{figure}
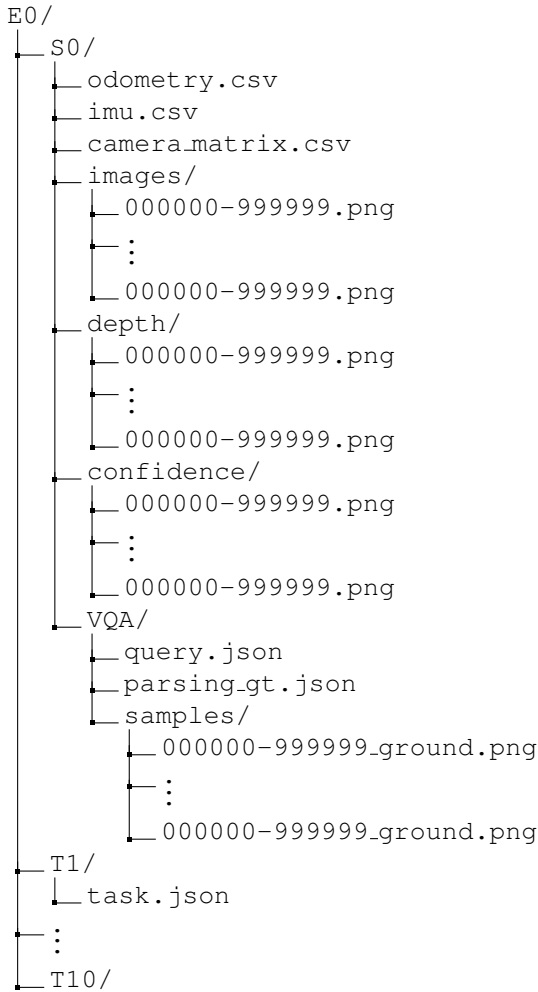

 \dirtree{%
 .1 E0/.
 .2 S0/.
 .3 odometry.csv.
 .3 imu.csv.
 .3 camera\_matrix.csv.
 .3 images/.
 .4 000000-999999.png.  
  .4 $\vdots$.
  .4 000000-999999.png.
  .3 depth/.
  .4 000000-999999.png.  
  .4 $\vdots$.
  .4 000000-999999.png.
  .3 confidence/.
  .4 000000-999999.png.  
  .4 $\vdots$.
  .4 000000-999999.png.
  .3 VQA/.      % Root directory (Level 1)
  .4 query.json.    
  .4 parsing\_gt.json.    
  .4 samples/.         % File inside src (Level 3)
  .5 000000-999999\_ground.png.  
  .5 $\vdots$.
  .5 000000-999999\_ground.png.
  .2 T1/.
   .3 task.json.
  .2 $\vdots$.
  .2 T10/.
}
    \label{fig:dirtree}
    \caption{\textbf{Dataset Structure.} Directory tree of an environment containing both scenarios and missions.}
\end{figure}

\section{Evaluation}
To demonstrate the merit of this dataset, we evaluate the performance of \method{}~\cite{zimmerman2026isrr} on multi-step visual sign grounding, single-step visual grounding and sign parsing. We choose \method{} as our baseline as it is the only approach that handles \textit{visual sign grounding} for non-trivial navigational signs and it also outperformed ~\cite{agrawal2026icraws} on the task of \textit{navigational sign understanding}. 

We use our \texttt{BenchmarkEngine.py} to run \method{} in a multi-step setup. 
We report the results in \tabref{tab:eval}.  \method{} achieves  over 84\% in sign parsing and over 52\% success rate on the task of visual sign grounding. However, it performs poorly on metrics related to long-horizon sign-based navigation that requires sequential decision making.

To address \textbf{Q1}, the local representation proposed by \method{} appears to be a promising candidate for visual sign grounding.
However, as \method{} is designed as a single-step visual sign grounding approach, it fails to complete many of the missions.
First, \method{} attempts to ground a location that is string-wise identical to the given goal. Therefore if the sign refers to a location ``COM1-COM3'' while the mission's goal is ``COM2'', \method{} will fail. This points to the importance of semantic reasoning over the sign's content and the user's instructions, in response to \textbf{Q2}. Pertaining to \textbf{Q4}, \method{} lacks a layer that can decompose complex, hierarchical human goal specifications to a sequence of subgoals, and therefore it fails to complete more complex mission goals such as ``Robot Living Studio, located in COM1''.

As \method{} is designed as a single-step visual sign grounding approach, it does not include any failure recovery mechanism. The evaluation shows that even when the grounding accuracy is high, such as 36/38 in the 4\textsuperscript{th} environment, it can still result in two failed missions. This emphasizes the importance of failure handling suggested by \textbf{Q3}. 
%To recover from failure, it is useful for the agent to have a memory of its past actions and observations, which relates to \textbf{Q2}.

%\paragraph{SignScene} 

% Authors are also encouraged to include benchmarks and/or leaderboards alongside their
% datasets to facility comparison of algorithms on specific use cases of the data.
% IJRR requires the utility of a new dataset to be demonstrated within the paper. In particular, a dataset should demonstrate how to use the data with an existing, publicly available, algorithm or technique. For example, if the paper is presenting a new dataset to be used for the development of visual SLAM algorithms, the paper should present the use of the dataset on an existing, ideally state-of-the-art, visual SLAM algorithm. Doing so gives potential users of the dataset confidence that the dataset is well prepared and that there are some challenging conditions in the dataset that an existing method cannot handle, thus spurring new research. Note that a dataset paper should not present a new algorithm in addition to the dataset. Datasets for which there are no existing algorithms to test on are not encouraged.

\begin{table*}[h]
\small\sf\centering
\caption{\textbf{Evaluation.} The performance of \method{} on the dataset, on the tasks of sign-based mapless navigation, visual sign grounding and sign parsing.\label{tab:eval}}
\begin{tabular}{c|c|c|c|c}
\toprule
Environment &  Success Rate (strict) & Success Rate & VQA Accuracy & Parsing Success  \\
\midrule
E1 & 2/13 & 2/13 &57/93 & 155/166\\ \midrule
E2 & 0/4 & 0/4 &14/26 & 85/103\\ \midrule
E3 & 2/10 & 2/10 &52/87 & 109/134\\ \midrule
E4 & 4/6 & 4/6 &32/34 & 79/90\\ \midrule
E5 & 1/7 & 1/7 &21/43 & 8/8\\ \midrule
E6 & 0/7 & 0/7 &23/48 & 63/73\\ \midrule
E7 & 1/9 & 1/9 &17/78 & 49/74\\ \midrule
total & 10/56 & 10/56 &216/409 & 548/648\\ 
\bottomrule
\end{tabular}
%\\[10pt]
\end{table*}

\section{Discussion} \label{sec:discussion}
We believe that this dataset will be useful to researchers working in a variety of areas related to embodied navigation in human-oriented environments. 
For the dataset, each scenario includes geometric, semantic and textual information. In addition we provide venue maps and room-level GPS-aligned scene graphs, both including the location of each scenario. Lastly, we provide ground truth parsing for each navigational sign. In addition to the original tasks of sign understanding, visual sign grounding and sign-based mapless navigation, we can imagine several other applications that can benefit from this dataset:
\paragraph{\textbf{Global Localization}} - using the observations from each scenario, a robot can localize itself using the floor plans, the scene graph or a combination of both. The observations can serve to construct a geometric and semantic local map that is matched against the floor plan or the scene graph, but also the signs themselves can be used for localization as demonstrated in \cite{zimmerman2026ral}. 
\paragraph{\textbf{Language-conditioned navigation}} - for complex tasks, such as ``Go toward the MRT station, take the escalator to level 2, then follow signs for exit B.'', the robot has to relate the language instruction, the observed signage, the scene graph and the venue map to reason about its next action.  

\paragraph{\textbf{Spatial-Semantic Reasoning}} - As the signs contain semantic information, they can be used to guide exploration tasks. Additionally, the consolidation of multiple priors can be evaluated through reasoning on sign-map consistency. Whether the observed sign agrees with the map can serve as a cue for localization, as a trigger for map update or as an uncertainty indicator for sign understanding. 

% Given that we provide the ground truth parsing for every sign, and a topological representation that captures both the structures (building, floors, room) and the navigational graph, the dataset can be utilized for planning long-horizon navigation from complex language specification. 
% \begin{itemize}
%     \item Exploration using semantic sign cues
%     \item Sign understanding
%     \item Sign grounding
%     \item Language-conditioned navigation (Go toward the MRT station, take the escalator to Level 2, then follow signs for Exit B.) The robot has to relate:
% linguistic instruction,
% observed signage,
% graph structure,
% map semantics
% \item Map–sign consistency reasoning - Does the observed sign agree with the provided map?
% \item Global localization on a scene graph
% \item Sign-assisted localization
%\end{itemize}

\section{Summary}
This paper introduced a dataset developed to enable sign-based visual navigation in human-oriented environments. However, the dataset has many other uses: the large quantity of scenarios collected in different public spaces, the venue maps and the scene graphs, make the dataset suitable for research in global localization, spatio-symbolic reasoning and language-conditioned navigation. In addition to providing the data in accessible, human-readable format, we also provide utility scripts for ROS 2 conversion in order to utilize various open-source packages from the ROS ecosystem. Scripts for visualization and data manipulation are provided in Python. 

% \section{Ethical Approval}
% The project is a systematic robotics/data-collection study, but it does not involve human subjects as defined by NUS-IRB\footnote{\url{https://nus.edu.sg/research/irb/resources/faq}}. Data are collected in publicly accessible environments using a handheld sensing device. Incidental bystanders may appear in the visual recordings, but researchers do not interact with them, collect information about them, track them, or analyze their behavior. The research concerns navigational signage and the spatial topology of the surrounding environment. No audio is recorded, and identifiable faces are anonymized prior to dataset release.

\section{Data Accessibility Statement}
The \dataset{} dataset and accompanying benchmark resources will be released through Hugging Face, and is available to download also in this link
% Prior to public release, the dataset will be made available to reviewers through a private, access-controlled link
\footnote{\url{https://drive.google.com/drive/folders/19mY5V9EvInMePndl342gCu6QuMp5SbQS?usp=sharing}}
. 
The release includes synchronized RGB, sparse depth, odometry, and IMU measurements, together with camera calibration parameters, visual grounding annotations, sign-parsing labels, mission definitions, sign-centric topological graphs, venue maps, and scenario-level spatial metadata.
%Because data were collected in publicly accessible environments, incidental bystanders may appear in the raw recordings. 
The public dataset will comply with applicable privacy, anonymization, and institutional requirements. We intend to distribute the dataset publicly on Hugging Face under a CC BY license. Dataset loaders, benchmark utilities, visualization tools, annotation-processing scripts, and ROS 2 conversion tools will be released alongside the data under the Apache License 2.0.

% \begin{acks}

% \end{acks}

% \begin{thebibliography}{99}

% \end{thebibliography}

\bibliographystyle{plainnat}
\bibliography{glorified, new}

\begin{thebibliography}{21}
\providecommand{\natexlab}[1]{#1}
\providecommand{\url}[1]{\texttt{#1}}
\expandafter\ifx\csname urlstyle\endcsname\relax
  \providecommand{\doi}[1]{doi: #1}\else
  \providecommand{\doi}{doi: \begingroup \urlstyle{rm}\Url}\fi

\bibitem[Agrawal et~al.(2026)Agrawal, Loo, Zimmerman, and Hsu]{agrawal2026icraws}
Ayush Agrawal, Joel Loo, Nicky Zimmerman, and David Hsu.
\newblock {Sign Language: Towards Sign Understanding for Robot Autonomy}.
\newblock In \emph{Multi-Modal Spatial AI for Robust Navigation and Open-World Understanding Workshop, IEEE Intl.~Conf.~on Robotics \& Automation (ICRA)}. IEEE, 2026.

\bibitem[Anderson et~al.(2018)Anderson, Wu, Teney, Bruce, Johnson, S{\"u}nderhauf, Reid, Gould, and Van Den~Hengel]{anderson2018cvpr}
Peter Anderson, Qi~Wu, Damien Teney, Jake Bruce, Mark Johnson, Niko S{\"u}nderhauf, Ian Reid, Stephen Gould, and Anton Van Den~Hengel.
\newblock Vision-and-language navigation: Interpreting visually-grounded navigation instructions in real environments.
\newblock In \emph{Proceedings of the IEEE conference on computer vision and pattern recognition}, pages 3674--3683, 2018.

\bibitem[{Apple Inc.}(2026)]{applearkit2026software}
{Apple Inc.}
\newblock \emph{ARKit}.
\newblock Apple Inc., Cupertino, California, USA, 2026.
\newblock URL \url{https://apple.com}.

\bibitem[Baruch et~al.(2021)Baruch, Chen, Dehghan, Feigin, Fu, Gebauer, Kurz, Dimry, Joffe, Schwartz, and Shulman]{baruch2021nips}
Gilad Baruch, Zhuoyuan Chen, Afshin Dehghan, Yuri Feigin, Peter Fu, Thomas Gebauer, Daniel Kurz, Tal Dimry, Brandon Joffe, Arik Schwartz, and Elad Shulman.
\newblock {ARK}itscenes: A diverse real-world dataset for 3d indoor scene understanding using mobile {RGB}-d data.
\newblock In \emph{Thirty-fifth Conference on Neural Information Processing Systems Datasets and Benchmarks Track (Round 1)}, 2021.

\bibitem[Cao et~al.(2026)Cao, Kumar, and Curtis]{cao2026arxiv}
Jing Cao, Nishanth Kumar, and Aidan Curtis.
\newblock Follow the signs: Using textual cues and llms to guide efficient robot navigation.
\newblock \emph{arXiv preprint arXiv:2601.06652}, 2026.

\bibitem[Chandaka et~al.(2025)Chandaka, Wang, Chen, Che, Zhai, and Wang]{chandaka2025corl}
Bhargav Chandaka, Gloria Wang, Haozhe Chen, Henry Che, Albert Zhai, and Shenlong Wang.
\newblock {Human-like Navigation in a World Built for Humans}.
\newblock In \emph{Conference on Robot Learning}, 2025.

\bibitem[Chen et~al.(2025)Chen, Lu, Yang, Zhang, Chen, Jia, and Pan]{chen2025ral}
Chang Chen, Liang Lu, Lei Yang, Yinqiang Zhang, Yizhou Chen, Ruixing Jia, and Jia Pan.
\newblock {Signage-Aware Exploration in Open World using Venue Maps}.
\newblock \emph{IEEE Robotics and Automation Letters (RA-L)}, 2025.

\bibitem[Gibson(2009)]{gibson2009pap}
David Gibson.
\newblock \emph{{The wayfinding handbook: Information design for public places}}.
\newblock Princeton Architectural Press, 2009.

\bibitem[Kim et~al.(2022)Kim, Kim, Song, Lee, Jung, and Kim]{kim2022sensors}
Pyojin Kim, Jungha Kim, Minkyeong Song, Yeoeun Lee, Moonkyeong Jung, and Hyeong-Geun Kim.
\newblock A benchmark comparison of four off-the-shelf proprietary visual--inertial odometry systems.
\newblock \emph{Sensors}, 22\penalty0 (24):\penalty0 9873, 2022.

\bibitem[Ku et~al.(2020)Ku, Anderson, Patel, Ie, and Baldridge]{ku2020emnlp}
Alexander Ku, Peter Anderson, Roma Patel, Eugene Ie, and Jason Baldridge.
\newblock Room-across-room: Multilingual vision-and-language navigation with dense spatiotemporal grounding.
\newblock In \emph{Proceedings of the 2020 Conference on Empirical Methods in Natural Language Processing (EMNLP)}, pages 4392--4412, 2020.

\bibitem[Liang et~al.(2020)Liang, Knepper, and Pokorny]{liang2020iros}
Claire Liang, Ross~A Knepper, and Florian~T Pokorny.
\newblock No map, no problem: A local sensing approach for navigation in human-made spaces using signs.
\newblock In \emph{Proc.~of the IEEE/RSJ Intl.~Conf.~on Intelligent Robots and Systems (IROS)}, pages 6148--6155. IEEE, 2020.

\bibitem[Lin et~al.(2025)Lin, Li, Zhao, Zhou, Wang, Wei, Tang, Li, Wang, Pang, et~al.]{lin2025arxiv}
Sihao Lin, Zerui Li, Xunyi Zhao, Gengze Zhou, Liuyi Wang, Rong Wei, Rui Tang, Juncheng Li, Hanqing Wang, Jiangmiao Pang, et~al.
\newblock Vlnverse: A benchmark for vision-language navigation with versatile, embodied, realistic simulation and evaluation.
\newblock \emph{arXiv preprint arXiv:2512.19021}, 2025.

\bibitem[{NVIDIA Corporation}(2026)]{nvidia2026software}
{NVIDIA Corporation}.
\newblock {NVIDIA Isaac Sim}: Robotics simulation platform.
\newblock \url{https://nvidia.com}, 2026.
\newblock URL \url{https://nvidia.com}.

\bibitem[Richter(2017)]{richter2017springer}
Kai-Florian Richter.
\newblock \emph{Indoor Wayfinding Tools}, pages 1035--1042.
\newblock Springer International Publishing, 2017.

\bibitem[Savva et~al.(2019)Savva, Kadian, Maksymets, Zhao, Wijmans, Jain, Straub, Liu, Koltun, Malik, Parikh, and Batra]{savva2019iccv}
Manolis Savva, Abhishek Kadian, Oleksandr Maksymets, Yili Zhao, Erik Wijmans, Bhavana Jain, Julian Straub, Jia Liu, Vladlen Koltun, Jitendra Malik, Devi Parikh, and Dhruv Batra.
\newblock Habitat: {A} {P}latform for {E}mbodied {AI} {R}esearch.
\newblock In \emph{Proceedings of the IEEE/CVF International Conference on Computer Vision (ICCV)}, 2019.

\bibitem[Sun et~al.(2026)Sun, Huang, Li, Xiao, Guo, Ghaffari, Li, Xu, and Kong]{sun2026arxiv}
Jian Sun, Yuming Huang, He~Li, Shuqi Xiao, Shenyan Guo, Maani Ghaffari, Qingbiao Li, Chengzhong Xu, and Hui Kong.
\newblock Signnav: Leveraging signage for semantic visual navigation in large-scale indoor environments.
\newblock \emph{arXiv preprint arXiv:2603.16166}, 2026.

\bibitem[Windecker et~al.(2025)Windecker, Patel, Reuss, Schwarzkopf, Cadena, Lioutikov, Hutter, and Frey]{windecker2025arxiv}
Tim Windecker, Manthan Patel, Moritz Reuss, Richard Schwarzkopf, Cesar Cadena, Rudolf Lioutikov, Marco Hutter, and Jonas Frey.
\newblock Navitrace: Evaluating embodied navigation of vision-language models.
\newblock \emph{arXiv preprint arXiv:2510.26909}, 2025.

\bibitem[Zhang et~al.(2025)Zhang, Sikchi, Zhang, and Biswas]{zhang2025rss}
Arthur Zhang, Harshit Sikchi, Amy Zhang, and Joydeep Biswas.
\newblock Creste: Scalable mapless navigation with internet scale priors and counterfactual guidance.
\newblock In \emph{Proceedings of Robotics: Science and Systems (RSS)}, 2025.
\newblock \doi{10.15607/RSS.2025.XXI.136}.

\bibitem[Zhao et~al.(2023)Zhao, Chen, Wang, and Lu]{zhao2023cvpr}
Haojie Zhao, Junsong Chen, Lijun Wang, and Huchuan Lu.
\newblock Arkittrack: A new diverse dataset for tracking using mobile rgb-d data.
\newblock \emph{CVPR}, 2023.

\bibitem[Zimmerman et~al.(2026{\natexlab{a}})Zimmerman, Loo, Agrawal, and Hsu]{zimmerman2026ral}
Nicky Zimmerman, Joel Loo, Ayush Agrawal, and David Hsu.
\newblock Signloc: Robust localization using navigation signs and public maps.
\newblock \emph{IEEE Robotics and Automation Letters}, 2026{\natexlab{a}}.

\bibitem[Zimmerman et~al.(2026{\natexlab{b}})Zimmerman, Loo, Koh, Wang, and Hsu]{zimmerman2026isrr}
Nicky Zimmerman, Joel Loo, Benjamin Koh, Zishuo Wang, and David Hsu.
\newblock {SignScene: Visual Sign Grounding for Mapless Navigation}.
\newblock In \emph{Proc.~of the Intl.~Symposium~on Robotic Research (ISRR)}, 2026{\natexlab{b}}.

\end{thebibliography}

\end{document}